\documentclass[11pt]{article}

\usepackage[margin=1in]{geometry}
\usepackage{amsmath,amssymb}
\usepackage[dvipsnames]{xcolor}
\usepackage{hyperref}
\usepackage{microtype}

\usepackage[hang,flushmargin]{footmisc}

\hypersetup{
    colorlinks=true,
    linkcolor=green!40!black,
    citecolor=green!40!black,
    urlcolor=green!40!black
}

\title{\Large\bf The Presupposition Problem in Representation Genesis}

\author{Yiling Wu\\
\normalsize BridgeM, Inc.}

\date{}

\begin{document}

\maketitle
\thispagestyle{empty}

\begin{abstract}
\noindent Large language models are the first systems to achieve high cognitive performance without clearly undergoing representation genesis: the transition from a non-representing physical system to one whose states guide behavior in a content-sensitive way. Every prior cognitive system we had studied had already made this transition before we could examine it, and philosophy of mind treated genesis as a background condition rather than an explanatory target. LLMs provide a case that does not clearly involve this transition, and this makes the genesis question urgent in a new way: if genesis did not occur, which cognitive capacities are affected, and why? We do not currently have the conceptual resources to answer this. The reason, this paper argues, is structural. The major frameworks in philosophy of mind, including the Language of Thought hypothesis, teleosemantics, predictive processing, enactivism, and genetic phenomenology, share a common feature when applied to the genesis question: each deploys, at some explanatory step, concepts whose purchase depends on the system already being organized as a representer. This pattern, which we call the Representation Presupposition structure, generates systematic explanatory deferral. Any attempt to explain the first acquisition of content-manipulable representation within the existing categorical vocabulary imports resources from the representational side of the very transition it is meant to explain. We call this the Representation Regress. The diagnostic result is that current philosophy of mind lacks the conceptual resources to explain representation genesis without presupposing what is to be explained. The paper is a conceptual diagnosis: it does not offer a new theory, but establishes the precise shape of the problem and derives two minimum adequacy conditions on any theory that avoids this structure. The appearance of LLMs is what makes the absence of a theory capable of explaining representation genesis consequential rather than merely theoretical.

\vspace{0.5em}
\end{abstract}

\section{Introduction}

Representation is among the most foundational concepts in the philosophy of mind and cognitive science. Theories of intentionality explain what representations are about, how they acquire their content, and how they guide behavior. Developmental and evolutionary accounts explain how representational repertoires expand and are transmitted across generations. Yet one question has received comparatively little systematic attention: how does any physical system first come to have representational states at all? This paper calls that question the representation genesis problem, and argues that the main philosophical frameworks considered here face a recurring structural difficulty when they are asked to address it.

The paper is a negative argument, and more specifically a conceptual diagnosis. Its goal is not to propose a new theory of representation, nor to argue that existing theories are wrong within their proper explanatory domains. Its goal is to show that the main frameworks considered here, when their explanatory structure is examined with respect to the genesis question, share a structural feature that generates systematic explanatory deferral when they are asked to answer it. We call this feature the Representation Presupposition structure. The argument for its presence across several major traditions constitutes the paper's first main contribution. From that diagnosis, the paper derives a Representation Regress: this is the name for the pattern in which any attempt to explain representation genesis using existing first-order vocabulary will import resources from the target side of the transition under investigation, generating systematic explanatory deferral. The regress motivates, as a second and more tentative contribution, the proposal that genesis may require a higher-order inquiry into the adequacy of the conceptual categories available for this specific explanatory task.

The importance of the genesis problem can be stated briefly. Representation is the explanatory primitive of much of cognitive science: mental states are characterized as representations, cognitive processes are characterized as operations over representations, and learning is characterized as the acquisition and modification of representations. If the genesis of content-manipulable representation cannot be explained within the existing vocabulary without systematic explanatory deferral, then a significant portion of cognitive-scientific theorizing rests on a presuppositional base that has not been examined. This does not render such theorizing wrong, but it means that its foundations are more opaque than typically acknowledged.

A methodological note must be made at the outset. The term "content-manipulable representation" is used throughout to pick out the target phenomenon. By this term we do not mean any state that might, under some liberal theory of representation, qualify as representational: we mean, in particular, states that support structure-preserving operations, including rule-governed inference and conceptual composition, and that are subject to misrepresentation. The argument is not that this is the only defensible notion of representation. It is that this is the phenomenon that the major frameworks under examination are typically invoked to explain, and that the genesis of this phenomenon in particular faces the structural difficulty the paper identifies.

A second methodological note concerns the vocabulary of transition. We resist the claim that the existing philosophical repertoire contains no concepts for describing the transition from non-representation to representation. Philosophers have proposed emergence, proto-content, structural coupling, developmental scaffolding, and dynamical organization as relevant notions. The paper's claim is more cautious: that when these and related concepts are examined at the point where they are invoked to explain first acquisition, they typically carry implicit conditions that are satisfied only by systems already organized as cognizers. The problem is not absence of vocabulary but the way in which the available vocabulary recurrently imports cognition-side commitments.

A third methodological note addresses a misreading that must be blocked at the outset. The paper might appear to proceed by first defining content-manipulable representation in a way that requires sophisticated cognitive capacities, and then noting that existing theories cannot explain how systems come to have such capacities. If this were the procedure, the paper would be manufacturing a gap rather than identifying one. The actual procedure is different. The target notion is not stipulated by the paper; it is identified as the kind of state that the major theories of cognition are themselves committed to explaining. LOT explains cognitive processes as computations over content-bearing mental symbols. Teleosemantics explains what gives a state its representational content. Predictive processing, in its representationally committed reading, attributes to cognitive systems generative models whose states bear structured expectations about the world. The genesis problem is not a demand that these theories explain something they were never designed to explain. It is a demand that they explain how systems come to have the very kind of states they take as their explanatory starting point. The gap is internal to the ambitions of these theories, not imposed on them from outside.

\section{The Target: Information, Encoding, and the Genesis Problem}

\subsection{Three Levels of Representational Vocabulary}

Before the argument can proceed, three notions must be distinguished. Conflating them generates the most persistent misunderstandings about what the genesis problem concerns and about what kinds of evidence or theoretical move would bear on it.

Natural information is a matter of nomological covariation. Smoke carries information about fire because, under the relevant conditions, the presence of smoke is reliably correlated with the presence of fire\footnote{Dretske, F. (1981). Knowledge and the Flow of Information. MIT Press. Dretske, F. (1988). Explaining Behavior: Reasons in a World of Causes. MIT Press, pp. 52-63. Dretske explicitly distinguishes natural information (nomological covariation) from representation (a state that plays a content-bearing role within a system's processing). His project of building representation from information is relevant here not as a refutation but as the most sustained attempt to close the gap: the difficulty his account faces at the learning-period step illustrates the structural problem this paper identifies.}. Tree rings carry information about a tree's age. In Shannon's sense, any variable Y carries information about X to the degree that knowing Y reduces uncertainty about X\footnote{Shannon, C. E. (1948). A mathematical theory of communication. Bell System Technical Journal, 27(3), 379-423. Shannon information H(X) = -sum p(x) log p(x) quantifies uncertainty reduction independently of semantic content. Mutual information I(X;Y) measures statistical dependence. Neither concept entails content-manipulable representation: a system can exhibit high mutual information with an environmental variable without having any state that represents that variable in the sense of playing a content-sensitive functional role.}. Natural information relations are pervasive in the physical world; they require no cognitive system and no intentional organization.

Statistical encoding is richer than natural information. A state is a statistical encoding if it compresses and represents distributional regularities in data: embedding vectors, weight matrices, and hidden-state activations in neural networks are examples. Statistical encodings can be generated by learning processes without presupposing prior content-manipulable representations, and they support operations that respect the distributional geometry of the training domain. Many theoretical traditions treat statistical encodings as a form of representation. The present paper does not contest this: in various theoretical frameworks, states of this kind may legitimately be characterized as representations. The point that matters here is different.

The target phenomenon of this paper is content-manipulable representation. This paper does not legislate the one true meaning of representation. It isolates a target kind of state that many leading theories of cognition are implicitly or explicitly committed to explaining, and it asks how any physical system first comes to have states of this kind.

Three core marks characterize the target. First, target-directedness: the state stands for some feature of the environment or of the system's own states in a way that is individuated by what it is directed at, not merely by the physical properties of the state token. Second, content-sensitive downstream use: the system's operations over the state are sensitive to its content, that is, to what it is about, not merely to its position within a distributional or syntactic space. Third, the possibility of system-level correctness and incorrectness: the state can misrepresent\footnote{Piccinini, G. (2020). Neurocognitive Mechanisms: Explaining Biological Cognition. Oxford University Press, ch. 3. A state has representational status if it plays a functional role involving standing in for its target within the system's processing. Importantly, in Piccinini's mechanistic framework, representing is itself a functional role defined at a mechanistic level of description; what this paper calls the genesis problem asks how that functional role is first instantiated, which requires an account that does not already presuppose a mechanism organized to instantiate it.}, which means it can be tokened in the absence of the feature it represents, and this constitutes a failure condition recognized within the system's own organization rather than merely an external observer's judgment of mismatch.

A clarification is needed before proceeding. The three marks defined here do not constitute a proposed universal definition of representation; they are not intended to rule out weaker notions that figure legitimately in other theoretical contexts. Their purpose is to delimit the explanatory target of this paper: the kind of state whose genesis faces the structural difficulty to be identified. Systems may have states that fall under weaker notions of representation without having states that meet these marks, and the paper takes no position on whether such states deserve the name representation in other contexts.

These three marks are core in the sense that a state lacking any of them is not a content-manipulable representation in the relevant sense. They should be distinguished from the paradigmatic manifestations that such states characteristically support: rule-governed inference, compositional recombination of representational elements, and reasoning under hypothetical or counterfactual assumptions. These manifestations are explanatory marks of the target phenomenon rather than additional defining conditions. A theorist who holds that misrepresentation is possible without full compositional productivity, or that content-sensitivity precedes sophisticated inference, is not thereby denying the target; rule-governed inference and composition are symptoms of the kind of state the core marks define, not additional requirements imposed by definitional fiat.

The distinction between the core marks and the paradigmatic manifestations matters for a specific reason. Critics may argue that the paper inflates the target by including sophisticated cognitive achievements in the definition of content-manipulable representation and then charging existing theories with failing to explain something they never aimed to explain. This objection is addressed directly in Section 7. Here it suffices to note: the core marks do not stipulate sophisticated inference or compositionality. They identify what content-sensitivity, target-directedness, and misrepresentation require of a state. The paradigmatic manifestations are listed as the kinds of behavior that such states characteristically support, not as threshold conditions a system must clear before its states qualify.

The working notion is theory-neutral enough to encompass the mental symbols of the Language of Thought hypothesis, the biological representations of teleosemantics, the generative models of predictive processing, and the perceptual representations of Burgean naturalism, while marking out a target distinctly richer than natural information or statistical encoding. The point of the three-level distinction is to identify the level whose genesis faces the structural difficulty the paper argues for, not to stipulate that only this level deserves the name representation. This characterization is consistent with representational realist and deflationary interpretations alike\footnote{Ramsey, W. M. (2007). Representation Reconsidered. Cambridge University Press. Ramsey distinguishes job descriptions for representational states and argues that most purported representations in cognitive science fail to satisfy genuinely representational criteria. His analysis of what it takes for a state to be a genuine representation, as opposed to a state that merely plays a functional role in processing, bears directly on the question of what content-manipulable representation requires. The characterization of the target in terms of target-directedness, content-sensitive downstream use, and misrepresentation possibility is aligned with the kind of criterion Ramsey identifies as genuinely representational rather than deflationary.}\footnote{Egan, F. (2014). How to think about mental content. Philosophical Studies, 170(1), 115-135. Egan argues that mental content attributions are theoretical posits rather than descriptions of intrinsic properties of mental states. On this view, representational content is an interpretive projection rather than a feature that physical states possess independently of the theorist. The genesis problem, as understood in this paper, arises under both realist and deflationary interpretations: even if content attribution is interpretive, the question of what makes a physical system eligible for the relevant kind of interpretive attribution remains open, and this eligibility question has the same presupposition structure the paper identifies.}: even under a deflationary reading of content attribution, the question of what makes a physical system eligible for the relevant interpretive treatment has the same presupposition structure the paper identifies, as Section 3 shows.

The contrast with statistical encoding is philosophically important and must be made explicit. A statistical encoding is tuned to distributional regularities in a training environment: it compresses those regularities and supports operations that respect the distributional geometry of the domain. A content-manipulable representation is individuated by what it represents rather than by where it falls in a statistical space. The difference is not one of complexity or sophistication. A highly complex statistical encoding remains a statistical encoding if its operations track distributional position rather than content; a relatively simple state can be a content-manipulable representation if its downstream use is genuinely sensitive to what it is about. Statistical encodings can be produced without prior content-manipulable representations, and they are genuinely useful. The question of genesis is whether the process that produces them also produces content-manipulable representation, or whether something more is required for the transition to the representational side.

\subsection{Natural Information and the Genesis Problem}

One might argue that the genesis problem dissolves if natural information relations are already present in the physical world: since smoke already carries information about fire, representation might emerge from such relations without any special genesis event. This objection must be addressed directly.

The objection moves too quickly between levels. Natural information is a matter of nomological covariation and requires no cognitive system. Content-manipulable representation requires a system that uses a state in ways sensitive to what it is about. The transition from a physical system that merely instantiates a state that covaries with some environmental feature to a system that uses that state as representing that feature is precisely the transition the genesis problem asks about. The presence of natural information relations in the world shows that informational resources are available; it does not explain how a system comes to exploit them representationally.

Dretske's project is relevant here as the most sustained attempt to close this gap rather than as an account that should simply be dismissed. Dretske explicitly distinguished natural information from representation and attempted to explain how indicator functions become representations through a learning period. The difficulty his account faces at exactly the learning-period step, where the system's acquisition of the function of responding to the indicator presupposes the relevant causal-functional architecture, illustrates the structural problem the paper identifies in a particularly transparent way. Dretske is, as it were, a witness to the difficulty rather than a solution to it.

\subsection{Burge's Treatment of Representation Genesis}

The most systematic recent philosophical treatment of the origin of perceptual representation is Tyler Burge's Origins of Objectivity (2010)\footnote{Burge, T. (2010). Origins of Objectivity. Oxford University Press. Burge's work is the most systematic recent philosophical treatment of perceptual representation genesis. He explicitly acknowledges that representation genesis is a distinct explanatory problem (pp. 1-30) and traces how perceptual systems come to represent objective particulars through a process of objectification. The observation in the text is that Burge's objectification process begins with sensory systems already capable of forming perceptual states with the right causal-functional connections: it explains the transition from proto-representational to objective representation rather than from completely non-representational physical organization to any representation at all. This is noted not to criticize Burge's project, which is explicit about its own scope, but to indicate where the present explanandum begins.}. Burge argues that perception constitutively involves representation of objective particulars and traces how perceptual systems come to represent mind-independent entities through a process he calls objectification. His work is important here for two reasons.

First, Burge is a genuine ally on the point that representation genesis is a real explanatory problem distinct from content determination, and his naturalistic-functional framework for addressing it sets a high standard for what such an account should look like. The diagnosis in Section 3 is not directed against his project's goals.

Second, and relevant to Section 3, Burge's objectification process begins with sensory systems already capable of forming perceptual states with the right causal-functional connections to the environment. What the process explains is the transition from proto-representational sensory states to objective representation of particulars: it is not an account of how a completely non-representational physical system first comes to have any states eligible for the process of objectification to operate on. Burge is explicit about the scope of his account. The point here is simply to register where the present explanandum begins relative to Burge's framework, not to criticize a project that is transparent about its own presuppositions.

\subsection{Delimiting the Genesis Problem}

Representation genesis, as used in this paper, is the process by which a physical system first comes to instantiate content-manipulable representational states. The genesis problem is the question of what this process is and which conceptual resources are adequate to describe it.

The genesis problem is distinct from four related problems that have received sustained philosophical attention. The content determination problem asks, given that a state represents, what determines which feature it represents. The normativity problem asks, given that a state represents, what explains the possibility of misrepresentation. The naturalization problem asks, given that a state represents, how its representational property is related to natural properties. The concept learning problem asks, given that a system already has content-manipulable representations, how it acquires new ones. Each of these presupposes that the system under analysis already has content-manipulable representational states. The genesis problem asks how any system comes to have such states in the first place, and this makes it prior to each of the others in a specific explanatory sense.

The genesis problem is also distinct from the Symbol Grounding Problem as formulated by Harnad (1990)\footnote{Harnad, S. (1990). The symbol grounding problem. Physica D: Nonlinear Phenomena, 42(1-3), 335-346. Grounding is a vertical dependence problem (how can symbols be grounded in non-symbolic states?). Genesis is a horizontal emergence problem (how does any system first come to have representational states?). A theory can solve grounding while leaving genesis untouched.}. Grounding is a vertical dependence problem: given a system with symbolic states, how can those symbols be grounded in non-symbolic states? Genesis is a horizontal emergence problem: how does a system first come to have states on the representational side at all? These are related but distinct problems.

\section{The Representation Presupposition Thesis}

We state the Representation Presupposition Thesis for the frameworks examined in this paper:

\noindent\textbf{RPT:} In the major frameworks examined here, when a framework is applied to the question of representation genesis, it deploys, at some identifiable explanatory step, at least one concept whose explanatory role is to organize a system that already has content-manipulable representational structure.

Four notions must be distinguished before the RPT can be stated with the precision it requires. Ordinary incompleteness: every theory has unexplained primitives; a theory is ordinarily incomplete when it does not explain some phenomenon that falls outside its intended domain. Scope limitation: a theory has a scope limitation when it explicitly restricts itself to a proper subset of the phenomena one might want explained. Explanatory dependence: a theory exhibits explanatory dependence when its explanatory success requires that certain conditions obtain, where those conditions are themselves left unexplained by the theory. Vicious presupposition: a theory exhibits vicious presupposition in the genesis context when the conditions on which its explanatory machinery depends themselves belong to the explanandum-side of the transition the theory is being asked to explain.\footnote{Levine, J. (1983). Materialism and qualia: The explanatory gap. Pacific Philosophical Quarterly, 64(4), 354-361. Levine identifies a structural asymmetry between functional-physical reduction and qualitative character: the explanatory gap arises when the explanandum cannot be derived from the explanans without residue. The Representation Presupposition structure identified in this paper is a related but distinct form of explanatory shortfall: not a gap between physical description and phenomenal character, but a failure of the explanatory starting point to stand on the non-representational side of the transition under investigation.}\footnote{Hempel, C. G., and Oppenheim, P. (1948). Studies in the logic of explanation. Philosophy of Science, 15(2), 135-175. The deductive-nomological model requires that the explanandum be derivable from the explanans together with laws. The circularity diagnosed in this paper is of a different structure: not that the explanandum is used as a premise, but that the explanatory starting point is drawn from the same side of the transition that the explanandum occupies, preventing non-trivial derivation of the first crossing.} The RPT identifies vicious presupposition, not ordinary incompleteness or scope limitation. A theory that is merely incomplete or scope-limited for genesis is not the target of the diagnosis. The target is a theory whose explanatory starting point, when pressed into genesis service, already describes a system as being on the representational side of the very transition whose first crossing is what genesis must explain.

We call a concept a cognition-side concept (CSIDE) in a genesis explanatory context when it plays an explanatory role such that its application to the system being characterized presupposes that the system already instantiates organizational features belonging to the representational side of the target transition. We use explanatory role rather than applicability conditions to be precise: the claim is not that CSIDE concepts are logically impossible to apply to non-representational systems, but that when they are deployed to explain genesis, their explanatory purchase depends on the system already having the kind of content-manipulable organization whose first appearance is what genesis is meant to explain. CSIDE status is therefore relational: a concept that is not CSIDE in its ordinary use as a description of an already-functioning cognitive system may become CSIDE when it is imported into a genesis explanation as a grounding primitive.

A word about the character of the diagnosis. The RPT does not charge these frameworks with logical error or internal inconsistency. The claim is not that they reason badly or that their arguments are invalid. The claim is about where they begin: their explanatory starting points, when examined relative to the genesis question, already describe systems that are on the representational side of the transition whose first crossing is what genesis must explain. This is an explanatory-level observation about the categorical structure of these frameworks, not an accusation of fallacious reasoning.

The RPT is a claim about the frameworks examined here, not an a priori claim about all possible theories. It is open to refutation by the identification of a framework that does not have this structure when applied to genesis. Section 7 discusses what a framework would have to look like to escape the structure. The analytical template applied to each framework is: (i) what is the framework designed to explain; (ii) what does it take as its explanatory starting point; (iii) does that starting point play a CSIDE explanatory role in the genesis context; and (iv) at which specific explanatory step does this become apparent. Crucially, the diagnosis does not say these frameworks are wrong in their proper domains. It says they were not designed to answer the genesis question, and when they are asked to do so, their explanatory primitives reveal a presupposition structure.

\subsection{The Language of Thought Hypothesis}

The Language of Thought hypothesis explains propositional attitudes and the computational processes that operate over them: mental processes are computations over syntactically structured representations whose semantic properties are fixed by causal-informational relations to the world. When the hypothesis is examined with respect to the genesis question, its explanatory structure becomes visible in a specific way.

The theory begins with a computational architecture over mental symbols\footnote{Fodor, J. A. (1975). The Language of Thought. Harvard University Press, ch. 2, pp. 55-97.}. Within that architecture, it explains how content is structured, combined, and put to inferential use. What it does not explain, and was not designed to explain, is how that architecture itself first comes to be. The language of thought, with its mental symbols, syntactic structure, and causal-informational relations, presupposes a system already organized as a content-manipulable representational system. It describes where genesis has already arrived, not how it got there.

Fodor's argument against concept learning makes the presupposition explicit\footnote{Fodor, J. A. (1981). Representations: Philosophical Essays on the Foundations of Cognitive Science. MIT Press, ch. 10, pp. 257-316.}\footnote{Fodor, J. A. (1998). Concepts: Where Cognitive Science Went Wrong. Oxford University Press, pp. 122-135.}. If acquiring concept C proceeds by hypothesis-testing, then C must already figure in the hypothesis space; no hypothesis-driven process can therefore explain first possession of C. Fodor's conclusion is that primitive concepts are triggered rather than learned\footnote{Fodor, J. A. (2008). LOT 2: The Language of Thought Revisited. Oxford University Press, ch. 6.}. For present purposes, what matters is not whether this conclusion is correct but what it reveals: the explanatory apparatus of the Language of Thought hypothesis does not reach first possession of representational structure. Fodor is, in this respect, a witness to the presupposition structure rather than a theorist to be refuted.

A point about statistical learning: one might object that deep learning systems generate statistical encodings from distributional learning without presupposing prior content-manipulable representations, which shows that the Fodor-style argument does not generalize. The objection is correct about statistical encodings, as Section 2.1 acknowledges. But the genesis problem, as characterized here, concerns the first acquisition of content-manipulable representation, not the generation of statistical encodings. Whether those encodings are themselves content-manipulable in the relevant sense is the question, not a premiss that can be assumed in response to the genesis problem.

\subsection{Teleosemantics}

Teleosemantics, developed by Millikan and elaborated by Papineau and Neander, explains what determines the content of a representational state\footnote{Millikan, R. G. (1984). Language, Thought, and Other Biological Categories. MIT Press, pp. 86-100.}\footnote{Papineau, D. (1987). Reality and Representation. Blackwell. Papineau, D. (1993). Philosophical Naturalism. Blackwell, ch. 3. Papineau's teleosemantic account, like Millikan's, explains why an existing representational state has the content it has. The contrast drawn in the text is not between teleosemantics and the target problem but between what teleosemantics was designed to do and what genesis explanation requires.}\footnote{Neander, K. (2017). A Mark of the Mental: In Defense of Informational Teleosemantics. MIT Press. Conditions for content-attribution within an already-functioning system.}. Its explanatory target is content determination within an already organized system, not the genesis of such organization. The examination here is not directed at teleosemantics in its proper domain; the question is what happens when it is pressed into genesis service.

The framework begins with mechanisms that have proper biological functions: mechanisms whose states have been selected because they systematically guided behavior relevant to the organism's reproductive success. A teleosemantic theorist will correctly point out that proper biological function is a naturalistic notion that does not require prior representational interpretation. It requires only that state-producing mechanisms have a selectional history of the appropriate kind. The diagnosis does not contest this point. The presupposition enters elsewhere.

Where the vicious presupposition enters: consider what it means for a state-type to be eligible for teleosemantic description. Not every physical state that varies with environmental features counts as a representation under teleosemantics; what matters is whether the state-producing mechanism has the right kind of selectional history and the right kind of downstream use. A system whose states are selected because they reliably guide behavior in the relevant way is one in which those states are already embedded in a downstream functional organization, one that uses them as guiding representations. Teleosemantics explains what determines the content of states within such an organization\footnote{Millikan, R. G. (2004). Varieties of Meaning. MIT Press, ch. 1. Millikan explicitly frames teleosemantics as a synchronic account of content-attribution for states within an already-functioning selected architecture. The question of how such architectures originate is identified as beyond the theory's stated scope.}\footnote{Shea, N. (2018). Representation in Cognitive Science. Oxford University Press, chs. 2-4. Shea develops a theory of representation grounding content in natural information, functional role, and system architecture. Crucially, Shea distinguishes the vehicle of representation from the system architecture in which it is embedded and argues that content attribution requires the vehicle be situated within an appropriate architectural context. From the perspective of the genesis argument, this architectural requirement functions as a CSIDE condition: it presupposes that the system already has an architecture of the kind that makes vehicle-content attribution applicable, which is precisely what genesis must explain.}. But the question of how a lineage of organisms first came to have state-producing mechanisms embedded in the kind of downstream organization that makes them eligible for teleosemantic description is precisely the genesis question. When teleosemantics is extended to answer this question, the concept of a selected mechanism with the right downstream use plays a CSIDE explanatory role: its application to the pre-genesis system already presupposes the kind of content-using organization whose first appearance is what must be explained. The presupposition is not in the notion of natural selection, which is uncontroversially non-representational, but in the notion of the organized downstream use that makes state-production eligible for selectional content-fixing in the first place.

\subsection{Predictive Processing}

The predictive processing framework, in its representationally committed formulation, explains perception, cognition, and action through a single organizational principle: the minimization of prediction error relative to a generative model. The generative model encodes prior probabilities P(H) over hidden environmental states and likelihood functions P(D\textbar H) specifying how those states generate sensory evidence D. Before locating the presupposition structure, a preliminary distinction is necessary.

A preliminary distinction is necessary, because predictive processing is internally contested on the question of representation. It can be interpreted as a representationally committed theory, in which the generative model is a content-manipulable representational structure, or as a purely formal or dynamical framework given a non-representational reading\footnote{Clark, A. (2016). Surfing Uncertainty: Prediction, Action, and the Embodied Mind. Oxford University Press. Clark's embodied reading minimizes representationalist commitment. The dilemma in the text is directed at representationally committed readings; Clark's non-representational reading escapes the specific presupposition charge but, if successful, removes predictive processing from the problem-space of intentional cognition as traditionally understood.}\footnote{Gładziejewski, P. (2016). Predictive coding and representationalism. Synthese, 193(2), 559-582. Gładziejewski examines whether predictive coding is genuinely committed to representations in a robust sense or whether it can be given a deflationary reading. He argues that the representational reading is indispensable for explaining the distinctive cognitive achievements predictive coding is invoked to explain, and that non-representational readings undercut the explanatory ambitions of the framework. This supports the dilemma stated in Section 3.3: the non-representational reading of predictive processing escapes the presupposition charge but only by removing the framework from the explanatory domain of intentional cognition.}. The diagnosis in this section is directed at the representationally committed reading. The non-representational reading escapes the specific presupposition charge examined here, but if it succeeds, it does so at the cost of removing predictive processing from the problem-space of intentionality as traditionally understood. This generates a dilemma for the predictive processing theorist who wishes to address the genesis of intentional cognition: either the framework is representationally committed and faces the presupposition structure identified below, or it is genuinely non-representational and does not directly bear on the target phenomenon.

A crucial distinction must be made before locating the presupposition structure: there is a difference between a prior as a formal mathematical parameter and a prior under the representational reading. A purely formal prior is a probability distribution over a hypothesis space; it can be described without attributing content-sensitive organization to the system. Under the representational reading, however, the prior encodes a structured expectation about hidden environmental states, and the system's states bearing those expectations are themselves content-manipulable representational states. The diagnosis is directed at the representational reading of the prior, not at its formal mathematical characterization.

Where the vicious presupposition enters under the representational reading: for the generative model to function as the framework describes, the model's state-space must already carve the space of possible environmental conditions in ways that are content-sensitive, that is, in ways individuated by what those conditions are, not merely by their statistical co-occurrence profiles. A model that merely tracks distributional regularities without content-individuation is a statistical encoding, not a generative model in the representational sense. Under the representational reading, therefore, the prior is not a neutral starting point but a structured content-bearing state\footnote{Friston, K. (2010). The free-energy principle: A unified brain theory? Nature Reviews Neuroscience, 11(2), 127-136. The generative model encodes prior probabilities P(H) over hidden states and likelihood functions P(D\textbar H) generating sensory evidence D.}. This is a CSIDE concept in the genesis context: invoking it to explain how a physical system first acquires representational organization already presupposes a system whose model of the environment is organized in content-sensitive ways.

The specific step: the vicious presupposition enters at the first appeal to the representational prior. The regress from priors to hyperpriors, acknowledged explicitly by Hohwy, relocates rather than dissolves the question\footnote{Hohwy, J. (2013). The Predictive Mind. Oxford University Press, p. 47. Hohwy acknowledges that the framework assumes a prior probability distribution over hidden causes and that the regress from priors to hyperpriors relocates rather than resolves the question of where the original prior comes from.}. Each level of the predictive hierarchy invokes a system with a model already in place, organized in content-sensitive ways. The genesis question is how a physical system first comes to have any such model at all, and predictive processing in its representational formulation does not address this. The non-representational reading of predictive processing, if it succeeds, escapes the specific presupposition charge, but at the cost noted above: it removes predictive processing from the domain of intentionality as traditionally understood, which is a cost the theorist who wishes to explain intentional cognition cannot easily accept.

\subsection{Enactivism and the Transition to Higher Cognition}

Radical enactivism holds that basic cognition is genuinely representation-free, with organisms engaging environments through sensorimotor coupling without requiring the tokening of content-manipulable representational states\footnote{Hutto, D. D., and Myin, E. (2013). Radicalizing Enactivism: Basic Minds without Content. MIT Press, pp. 3-45 and 155-170. Hutto and Myin explicitly identify the transition to content-involving cognition as "the hard problem of content," acknowledging that this transition remains unexplained within their framework. This is used here as an acknowledgment by enactivists themselves that the transition is theoretically problematic.}\footnote{Thompson, E. (2007). Mind in Life: Biology, Phenomenology, and the Sciences of Mind. Harvard University Press, ch. 5.}. This appears to offer an account that avoids the presupposition problem for basic cognition, and on this point the argument does not contest the enactivist claim.

The paper's concern with enactivism is conditional. It does not claim that all cognition requires content-manipulable representation. It raises the following, more limited question: if certain forms of cognitive achievement, including reasoning about absent or abstract or counterfactually specified states of affairs, plausibly involve stable content-sensitive operations that do not reduce to sensorimotor coupling with currently present environments, then the transition to such achievements requires explanation. Whether any given cognitive achievement falls in this category is itself a substantive question the paper does not pretend to resolve. The conditional is offered as a diagnostic probe rather than as a settled thesis about the nature of higher cognition.

Hutto and Myin explicitly acknowledge that the transition to content-involving cognition is what they call the hard problem of content. This acknowledgment is significant: it indicates that the enactivist tradition itself recognizes the transition as theoretically problematic. The point is not that enactivism is wrong but that, at the point of transition to content-involving cognition, the same structural difficulty appears that appears in the other frameworks.

Clark and Toribio (1994) argue that some forms of higher cognition seem to require representational states that basic sensorimotor coupling cannot supply\footnote{Clark, A., and Toribio, J. (1994). Doing without representing? Synthese, 101(3), 401-431.}. Whether they are correct is not settled here; the observation is used only to indicate that even within the enactivist tradition, the representational question does not disappear at the level of higher cognition.

\subsection{Genetic Phenomenology}

Brentano introduced intentionality as the mark of the mental\footnote{Brentano, F. (1874/1995). Psychology from an Empirical Standpoint (A. C. Rancurello, D. B. Terrell, and L. L. McAlister, Trans.). Routledge. Book II, ch. 1.}. Husserl developed this into a systematic analysis of intentional structure, including in his genetic phenomenology an account of how intentional contents are constituted over time through passive synthesis\footnote{Husserl, E. (1939/1973). Experience and Judgment (J. S. Churchill and K. Ameriks, Trans.). Northwestern University Press. Husserl, E. (1966/2001). Analyses Concerning Passive and Active Synthesis (A. Steinbock, Trans.). Kluwer.}\footnote{Husserl, E. (1900-1901/1970). Logical Investigations (J. N. Findlay, Trans.). Routledge.}. The word "genesis" appears in Husserl's own vocabulary, making it important to be precise about the difference between his explanandum and the one addressed here.

Husserlian genetic analysis concerns the constitution of intentional contents within an already-given intentional field\footnote{Zahavi, D. (2003). Husserl's Phenomenology. Stanford University Press, ch. 4.}. Its starting point is a transcendental consciousness with pre-reflective intentional directedness, and it traces how determinate contents, object-constitution, temporal synthesis, familiarity, are built up from more primitive intentional processes. The transcendental ego with its intentional field is the condition of the analysis, not its explanandum.

The difference between the phenomenological starting point and the present explanandum is therefore not a criticism of Husserl but a specification of different problem-spaces. Phenomenological genesis explains constitution within an intentional field; the genesis problem addressed here concerns how any physical system comes to have an intentional field. These are different explanatory questions with different starting points, and phenomenology's first-person methodology makes it structurally unsuited for the second question, not because first-person descriptions are wrong, but because one cannot describe the genesis of a first-person perspective from within the first-person perspective without presupposing what is to be explained.

\subsection{The Recurring Structural Pattern}

Across the five frameworks, the specific explanatory primitive differs: mental symbol, selected mechanism, generative model with a representational prior, sensorimotor coupling transitioning to higher cognition, transcendental intentional field. The frameworks differ in ontology, in methodology, in their commitments about the nature of mental states, and in the level of description at which their explanations operate. What they share is a structural feature that becomes visible only when they are asked to address first acquisition rather than content-determination, mechanism specification, or cognitive organization within an already-functioning system.

The significance of the convergence deserves explicit statement, because it is the inferential bridge from the individual case analyses to the general thesis. The convergence is not doctrinal: the frameworks do not share a substantive theoretical commitment that could explain why they all end up with the same starting point. LOT and teleosemantics differ profoundly on the nature of mental content. Enactivism explicitly rejects the representationalism that LOT and predictive processing presuppose. Genetic phenomenology operates within a first-person methodology entirely different from the naturalistic frameworks. Despite these differences, each framework, when pressed into genesis service, reaches for an explanatory starting point that already describes the system as organized on the representational side of the transition. The convergence is therefore structural rather than doctrinal: it reflects not what these theories agree about, but where they all begin when asked to explain genesis. It is this structural convergence, across otherwise incompatible frameworks, that gives the Representation Presupposition Thesis its force as a general diagnosis rather than as a set of disconnected complaints about individual theories.

This is not a charge that the frameworks are wrong about what they aim to explain. It is a diagnosis of where they begin, relative to the genesis question. To see that this beginning point is not obviously the right one for genesis explanation is already to see that genesis is a distinct problem. The rest of the paper develops the consequences of this observation.

\section{The Representation Regress}

\subsection{From Presupposition to Regress}

The Representation Presupposition Thesis, as established for the frameworks examined here, generates what we call the Representation Regress. The regress does not claim that explanation is impossible in general, nor that the identified frameworks are internally inconsistent. Its claim is more specific: within the existing categorical vocabulary, attempts to explain the first acquisition of content-manipulable representation face a pattern of explanatory deferral without reaching a non-presupposing terminus.

Consider any proposed explanation E of how system S first acquires content-manipulable representational state R, framed within the existing vocabulary. By the RPT as applied to the main frameworks, E deploys at some step a concept C with a CSIDE explanatory role. This means that E presupposes some content-manipulable representational organization in S, call it R*, as a condition on the explanation's getting started. E thus does not explain first acquisition of content-manipulable representation from a non-representational starting point; it explains a representational transition within an already partially representational system.

We can now ask what explains how S acquired R*. Any explanation E* of this acquisition, framed within the existing vocabulary, will by the same structural pattern deploy a concept with a CSIDE explanatory role, presupposing yet further representational organization R**. This is the regress.

The regress is not defeated by appeal to innateness. If R* is innate, the question shifts to the evolutionary level: how did the lineage of S come to produce organisms with innate content-manipulable representational organization? Explanations at this level deploy concepts, including selected mechanism and adaptive relevance, that in the genesis context carry the same CSIDE explanatory role. The presupposition is relocated, not dissolved.

A crucial clarification must be made here, since it is the site of the most likely misreading. The regress does not prohibit appeals to physical description. A theorist may always say: stop at the physical; describe the relevant neural states, and we are done. The regress's reply to this move is not that physical description is forbidden but that it is not yet genesis explanation. To say that certain neural states exist is not yet to say that they are content-manipulable representations or that they play a representational role within the system. The moment one tries to explain why those physical states count as representations, or how they come to play a representational role, the existing vocabulary is invoked, and the pattern identified by the RPT recurs. Physical description provides the substrate; it does not, on its own, explain genesis.

\subsection{Formal Structure}

We offer a partial formalization. The formalization is not intended to replace the prose argument but to make explicit the inferential structure that the prose makes vivid. Four predicates are used:

\begin{center}\emph{EXP(E, S, R): explanation E accounts for how system S first acquires R}\end{center}

\begin{center}\emph{DEP(E, C): explanation E employs concept C in an explanatory role}\end{center}

\begin{center}\emph{CSIDE(E, C): concept C plays a cognition-side explanatory role in E}\end{center}

\begin{center}\emph{REP(O): organizational feature O belongs to the representational side}\end{center}

The CSIDE predicate requires an explicit characterization:

\begin{center}
\emph{CSIDE(E, C) iff:}\\
\emph{DEP(E, C), and}\\
\emph{C's explanatory purchase in E depends on the system S being characterized}\\
\emph{as already instantiating some organizational feature O such that REP(O),}\\
\emph{where O is not identical to R and is not derivable from non-REP features}\\
\emph{by the resources available to E.}
\end{center}

This characterization makes precise the sense in which CSIDE is not ordinary incompleteness. Ordinary incompleteness occurs when DEP(E, C) and C is unexplained, but C's explanatory purchase does not depend on the REP condition above. Scope limitation occurs when E does not attempt EXP(E, S, R) at all. Explanatory dependence occurs when E's success conditions require conditions that E leaves unexplained, without those conditions being REP. Vicious presupposition, which is what the RPT identifies, is the specific case where the required conditions are REP and the REP feature is relevantly connected to R itself.

The Representation Presupposition Thesis for the frameworks examined:

\begin{center}
\emph{RPT: For each framework F among \{LOT, Teleosemantics, PP, Enactivism,}\\
\emph{Genetic Phenomenology\}, restricted to representationally committed}\\
\emph{readings where applicable:}\\
\emph{if F is applied to yield EXP(E, S, R) for any genesis target R,}\\
\emph{then there exists C such that DEP(E, C) and CSIDE(E, C).}
\end{center}

The restriction to representationally committed readings is explicit in the formalization because enactivism and predictive processing each admit of non-representational readings that would not be subject to this diagnosis, as Section 3 acknowledges. The RPT as stated is a claim about the representationally committed versions of those frameworks, and about LOT, teleosemantics, and genetic phenomenology without qualification.

Define an explanation sequence for R in system S as a sequence (E\_1, E\_2, ...) where E\_1 is meant to explain first acquisition of R, and each subsequent E\_i is meant to explain the representational organization presupposed by E\_\{i-1\}. Formally, if CSIDE(E\_i, C\_i), then E\_\{i+1\} is meant to explain how S comes to instantiate the REP feature that C\_i's explanatory role requires. The regress claim:

\begin{center}
\emph{Regress: For any finite explanation sequence (E\_1, ..., E\_n) for R in S,}\\
\emph{constructed from the categorical vocabulary of the examined}\\
\emph{frameworks: there exists no terminal index n such that}\\
\emph{not-CSIDE(E\_n, C) for all C such that DEP(E\_n, C).}
\end{center}

The argument for the regress claim: by RPT applied at each step, every E\_i in the sequence deploys some C\_i with CSIDE(E\_i, C\_i). Any putative termination point E\_n either (a) appeals to a representational primitive, in which case CSIDE(E\_n, C\_n) holds by the RPT and E\_n is not genuinely terminal, or (b) appeals to a purely physical state description, in which case E\_n has not yet explained why that physical state instantiates any REP feature, and any explanation of this that uses the examined vocabulary will by RPT deploy some concept with a CSIDE role. The key point about (b): physical description provides a substrate characterization, not a genesis explanation. To move from physical state to representational status requires invoking the very vocabulary whose application in genesis contexts is subject to the RPT. Neither path to termination is available within the existing categorical vocabulary.

Three forms of CSIDE explanatory role can be distinguished. In possession presupposition, the explanation invokes a component of the system whose existence already constitutes having content-manipulable representational states. In functional presupposition, the explanation invokes a mechanism whose operation presupposes a system that already uses its outputs in ways that make them eligible for content-attribution. In deployment presupposition, the explanation invokes a process that operates over representational or proto-representational distinctions that are themselves the target of the explanation. The five frameworks examined in Section 3 can each be located primarily in one of these three forms.

\subsection{Explanatory Deferral and Its Implications}

The Representation Regress establishes that current categories are subject to explanatory deferral for first-acquisition explanation: any attempt to use them to describe the transition from a non-representing to a representing system will at some point import resources from the representational side of that very transition. The relevant notion is deferral rather than impossibility: the claim is not that genesis is inexplicable in principle, but that the existing vocabulary defers the problem rather than resolving it.

It is important to distinguish explanatory deferral from two things it might be confused with. It is not ordinary explanatory incompleteness: every theory has unexplained primitives, and that alone would not be distinctive. The relevant structural feature is that the primitives in the genesis context are not merely unexplained but are drawn from the explanandum-side: they describe precisely the kind of system whose first appearance is what genesis is meant to explain. And it is not a logical paradox or a formal contradiction: the frameworks are internally consistent, and they are not claiming to explain genesis in most cases. The issue is what happens when they are asked to do so.

\section{A Higher-Order Inquiry into Category Adequacy}

\subsection{What the Regress Motivates}

The Representation Regress shows that the categorical vocabulary of philosophy of mind seems insufficient, in its current form, to explain representation genesis without systematic explanatory deferral. The natural response to this diagnosis is not to conclude that genesis is inexplicable but to ask what kind of inquiry would be needed. This paper argues, provisionally, that it may require an inquiry of a higher order than ordinary first-order philosophy of mind: an investigation into which conceptual categories are adequate for describing the genesis transition without importing cognition-side presuppositions. We call this, for want of a better label, a higher-order inquiry into category adequacy for representation genesis. We use "meta-ontology of mind" as a provisional shorthand, while acknowledging that this label might suggest more architecturally developed commitments than the paper is able to defend.

The case for a distinct research level can be made by contrasting two types of inquiry. First-order philosophy of mind asks: what are representations, how do they acquire content, how do they relate to behavior, are they identical to physical states? It takes the category of representation as given and investigates what falls under it or what constitutes it. The inquiry motivated by the regress asks a different question: whether the available categories are adequate for describing the genesis transition without generating the deferral pattern. This is not a question about what falls under the category of representation but about whether the category and its associated theoretical apparatus are the right tools for a particular explanatory task.

The analogy to philosophy of physics is instructive, though it should not be pressed too far. Physics asks what laws govern physical systems. Philosophy of physics asks whether the categories of our best physical theories, space, time, particle, field, are adequate: whether they carve the relevant domain correctly, whether they introduce hidden presuppositions, whether a different categorical framework might do better. The question of category adequacy for a specific explanatory task is different from asking questions within the categories. Similarly, the adequacy of first-order philosophical concepts for describing genesis is a question that first-order theorizing cannot answer by itself, because first-order theorizing presupposes those concepts.

The Kantian framing is instructive in a restricted methodological sense: a higher-order inquiry into category adequacy asks which categories are necessary conditions for the possibility of genesis explanation that avoids systematic deferral, rather than asking what genesis is within the existing categories\footnote{Kant, I. (1781/1787/1998). Critique of Pure Reason (P. Guyer and A. W. Wood, Trans.). Cambridge University Press. A50-52/B74-76. The Kantian framing is used only in a restricted methodological sense: the higher-order inquiry asks which categories are necessary conditions for the possibility of representational theorizing about genesis. No commitment to transcendental idealism is implied.}. No commitment to transcendental idealism is implied. The analogy is methodological, not metaphysical.

\subsection{Minimum Adequacy Conditions}

From the Representation Regress, two minimum adequacy conditions can be derived for any theory of representation genesis that avoids the identified deferral pattern. These are stated as necessary conditions on any future theory rather than as components of an already-developed framework.

CC-1 (Non-Presupposition): A theory of representation genesis must employ no concepts that, in their explanatory role within that theory, presuppose the possession of content-manipulable representational states by the system being characterized in its pre-transitional state.

CC-2 (Transition Intelligibility): A theory of representation genesis must describe the transition from the pre-representational to the representational state in terms permitted by CC-1, in such a way that the transition is intelligible and not arbitrary from within the non-presupposing vocabulary.

CC-1 follows directly from the regress: any theory that violates it is subject to the deferral pattern. CC-2 follows from the requirement that an account of genesis must do more than identify endpoints; it must make the transition itself intelligible. Together, these conditions constitute what a theory would have to satisfy to escape the regress. The RPT, as applied to the frameworks examined here, shows that none currently satisfies CC-1.

Two supplementary methodological aspirations are worth noting alongside these necessary conditions, though they are not derived from the regress. SC-1 (Formal Expressibility): a theory adequate for the genre of scrutiny this problem invites would benefit from being expressible in a formal language precise enough to permit logical derivation of consequences. SC-2 (Empirical Coherence): such a theory should be broadly consistent with empirical findings in cognitive development, evolutionary biology, and artificial systems research. These are aspirations rather than necessary conditions, and the paper does not impose them as requirements any proposed theory must immediately satisfy.

\section{Large Language Models as Diagnostic Case}

The philosophical argument advanced in Sections 2-5 does not depend on any particular empirical verdict about current large language models. The case introduced here is a diagnostic illustration: it is meant to show that the distinction drawn in Section 2.1 between statistical encoding and content-manipulable representation corresponds to a difference that has observable consequences in a constructed system, and to illustrate what representation genesis is not. Large language models are theoretically significant for the genesis problem in a specific methodological sense: they constitute, for the first time, systems of enormous functional sophistication that have been produced without undergoing anything resembling the perceptual, developmental, or evolutionary processes through which biological representational systems are thought to arise. This makes them a useful comparison class: by examining where their capacities diverge from those of systems that have undergone genesis, we can make the distinction between statistical encoding and content-manipulable representation empirically tractable in a way that is otherwise difficult to achieve. The section does not attempt to prove that large language models lack content-manipulable representations, which would require more than a diagnostic case can provide. It offers a conceptual worked example of why statistical sophistication and linguistic fluency do not settle the genesis question.

Large language models are trained through distributional learning over large corpora, producing internal states that encode the statistical structure of those corpora with high fidelity. These states are statistical encodings in the sense of Section 2.1. The philosophical question is whether they are also content-manipulable representations in the sense that matters for the genesis discussion.

The diagnostic observation can be stated as a conditional principle:

\begin{center}
\emph{CMT: A system that plausibly has content-manipulable representations}\\
\emph{would be expected to perform structure-preserving operations}\\
\emph{that track content rather than distributional statistics}\\
\emph{even when distributional statistics and content come apart.}
\end{center}

CMT is stated as a diagnostic expectation rather than a necessary condition, because the paper does not require the strong biconditional form. The point is that if statistical encoding and content-manipulable representation come apart in the way Section 2.1 describes, we would expect to see performance asymmetries at exactly the points where distributional patterns and domain structure diverge. The evidence from the evaluation literature is broadly consistent with this expectation.

The evaluation literature converges on a consistent pattern. Wan et al. (2024) find that large language models are sensitive to surface lexical and ordering features in ways that track corpus statistics rather than logical structure\footnote{Wan, Y., Wang, W., Yang, Y., Yuan, Y., Huang, J. T., He, P., Li, H., and Lyu, M. R. (2024). A and B equals B and A: Triggering logical reasoning failures in large language models. arXiv:2401.00757.}; Zhou et al. (2024) find performance degradation when semantic content is decoupled from common-sense statistical associations, even when the logical structure of the task is held constant\footnote{Zhou, B., et al. (2024). Conceptual and unbiased reasoning in language models. arXiv:2404.00205.}. Bubeck et al. (2023), in a broad exploratory study, document a characteristic profile across tasks: strong performance where distributional regularities reliably cue the correct response, alongside fragility precisely where that cuing fails\footnote{Bubeck, S., et al. (2023). Sparks of artificial general intelligence: Early experiments with GPT-4. arXiv:2303.12712.}. The causal reasoning literature sharpens the diagnosis further: Pearl (2009) establishes that causal structure is not reducible to observational statistics, and distributional training is in principle insufficient to generate states that track causal rather than merely correlational structure\footnote{Pearl, J. (2009). Causality: Models, Reasoning, and Inference (2nd ed.). Cambridge University Press, ch. 1.}. Taken together, these findings are consistent with systems whose internal states are tuned to distributional regularities rather than to content in the sense the genesis discussion requires.

The philosophical significance of this diagnostic is not to adjudicate the empirical debate about LLM capabilities but to make the Section 2.1 distinction vivid: a system can achieve very sophisticated distributional competence, including linguistic behavior that mirrors human use in most respects, without having undergone representation genesis in the relevant sense\footnote{Mahowald, K., Ivanova, A. A., Blank, I. A., Kanwisher, N., Tenenbaum, J. B., and Fedorenko, E. (2024). Dissociating language and thought in large language models: A cognitive perspective. Trends in Cognitive Sciences, 28(6), 517-540.}\footnote{Fedorenko, E., Piantadosi, S. T., and Gibson, E. A. (2024). Language is primarily a tool for communication rather than thought. Nature, 630(8017), 575-586.}. This illustrates, at the level of a constructed system, why linguistic fluency does not settle the genesis question and why the distinction between statistical encoding and content-manipulable representation is not merely conceptual.

\section{Objections and Replies}

\subsection{The Stopping Point Objection}

A natural objection holds that all explanation must stop somewhere, and that the frameworks examined here simply stop at a point that has not yet been explained. Every theory has unexplained primitives; what the diagnosis identifies as a presupposition structure might be nothing more than the ordinary incompleteness shared by all theoretical accounts.

The objection correctly identifies the general phenomenon but misidentifies the specific structural feature at issue. There is a difference between a stopping point that is merely unexplained and one that is drawn from the explanandum-side of the very transition under investigation.

An ordinary unexplained primitive is a concept that is taken as given because its explanation belongs to a different inquiry or a more fundamental theory. The stopping point of classical genetics, the gene as a unit of heredity, was not itself the phenomenon genetics was explaining; it was a functional posit invoked to explain inheritance patterns, and its chemical explanation came from outside genetics. By contrast, a stopping point on the explanandum-side describes precisely the kind of system whose first appearance is what the explanation is meant to account for. Invoking a language of thought, a generative model, or a selected representational architecture to explain representation genesis is not the same as invoking the gene to explain heredity: it is more like invoking genetic replication to explain how the first replicator appeared.

The diagnosis of the frameworks examined is that their stopping points, in the genesis context, are of the second kind. This does not mean they are wrong to have those stopping points for their intended purposes. It means those stopping points are inadequate when the explanatory target is genesis.

\subsection{The Mechanistic Explanation Objection}

Mechanistic accounts of cognition, such as those developed by Piccinini, explain representational status by appeal to the physical organization of mechanisms rather than by invoking prior representational structure\footnote{Piccinini, G. (2020). Neurocognitive Mechanisms: Explaining Biological Cognition. Oxford University Press, ch. 3. A state has representational status if it plays a functional role involving standing in for its target within the system's processing. Importantly, in Piccinini's mechanistic framework, representing is itself a functional role defined at a mechanistic level of description; what this paper calls the genesis problem asks how that functional role is first instantiated, which requires an account that does not already presuppose a mechanism organized to instantiate it.}. Since mechanistic explanation begins with physical and functional relations among components, it might seem to offer a route to genesis explanation that does not face the presupposition structure identified in Section 3. Why cannot genesis be explained as a matter of increasingly organized mechanisms giving rise to representational states?

The mechanistic approach is the most promising naturalistic framework for addressing the genesis question, and the paper does not claim it cannot contribute to an answer. The point is more specific.

Mechanistic descriptions explain how states are produced, transformed, and stabilized within a system whose components stand in certain physical and functional relations\footnote{Bechtel, W. (2008). Mental Mechanisms: Philosophical Perspectives on Cognitive Neuroscience. Routledge, ch. 2. Bechtel develops an account of mechanistic explanation for cognitive science in which cognitive phenomena are explained by identifying the components, operations, and organization of the mechanisms that produce them. His account does not directly address the genesis problem as formulated here, but it provides the most developed version of mechanistic explanation for cognition and is the natural background for the mechanistic objection addressed in Section 7.2.}. A mechanistic explanation of representation would say: when a mechanism is organized in such-and-such a way, its states count as representations. The question the regress poses is: what makes those states count as representations rather than merely as physical states with a certain causal profile? The moment a mechanistic account tries to answer this question, it must invoke some notion of what it is for a state to represent, and that notion, if it is to be informative, will characterize the state in terms of its role within a system already organized to use it as representing something. At that point, the mechanistic account faces the same structural issue the other frameworks face. This is visible even in more refined mechanistic accounts: distinguishing structural representations from mere detectors requires invoking conditions on the downstream processing organization that presuppose the very kind of content-using architecture whose genesis must be explained\footnote{Gładziejewski, P., and Miłkowski, M. (2017). Structural representations: Causally relevant and different from detectors. Biology and Philosophy, 32(3), 337-355. The authors distinguish structural representations, whose content derives from a structural isomorphism between the representation and what it represents, from mere detectors, which respond reliably to environmental features without encoding their structure. This distinction is relevant to the CSIDE analysis of mechanistic accounts: a system with states that are mere detectors in this sense is not yet a system with content-manipulable representations, even if the detector states play a causal role in processing. Genesis, on this analysis, requires explaining the transition from detector-grade states to structurally representational states, and mechanistic accounts that explain the former do not automatically explain the latter.}.

This is not a refutation of mechanistic accounts; it is a specification of the explanatory step they must cross. A mechanistic theory of representation genesis that satisfies CC-1 and CC-2 would be a significant achievement. The regress argument says this achievement has not yet been secured by existing mechanistic frameworks, and that securing it would require the kind of higher-order inquiry into category adequacy that Section 5 motivates. Mechanistic accounts of the production and transformation of states are entirely compatible with this: what remains is to explain why mechanistic organization of the right kind constitutes genesis rather than merely describing its physical subvenience base.

\subsection{The Boundary Objection}

The term regress, as used in this paper, may suggest a stronger claim than is intended: that of an infinite series of explanatory steps, each generating a further demand. But what the analysis actually establishes is a finite explanatory boundary: the existing frameworks stop at a certain point and cannot go further. Is a finite boundary really a regress?

The term regress is used to describe a pattern of explanatory deferral rather than to make a claim about an actually infinite series. The point is that the deferral is systematic: for any explanation within the existing vocabulary that purports to explain genesis, there will be a step at which a cognition-side resource is imported, and that step generates a further explanatory demand that the same vocabulary cannot meet without importing a further cognition-side resource. "Deferral" might be a more neutral term, and readers who prefer it can substitute it throughout without loss of content. The substantive claim is about the systematic character of the pattern, not about its infinite extension.

\subsection{The Scope of the RPT}

A concern about scope: the RPT is established by examining five theoretical traditions. Even granting the case analyses, why should a generalization from five traditions bear on philosophy of mind as a whole? The diagnosis might establish a localized structural feature of a particular selection of frameworks rather than a general result.

Reply: The RPT is not stated as a claim about all of philosophy of mind; it is stated as a claim about the major frameworks examined here. The scope of the claim is exactly the scope of the examination. The significance of the claim is proportional to the representativeness of the frameworks examined: the Language of Thought hypothesis, teleosemantics, predictive processing, enactivism, and genetic phenomenology between them cover a very substantial portion of the contemporary theoretical landscape. But the paper explicitly invites refutation by counter-example: a framework that satisfies CC-1 would constitute a response to the diagnosis. The analytical template of Section 3 provides the tool for evaluating any proposed counter-example.

\subsection{The Target Inflation Objection}

A further concern targets the paper's characterization of the explanandum. By defining content-manipulable representation in terms that include target-directedness, content-sensitive downstream use, and the possibility of misrepresentation, the paper might be seen as inflating the target: charging existing theories with failing to explain something they were never designed to address, and manufacturing a gap that would dissolve under a more modest notion of representation.

Reply: The objection rests on a misreading of the paper's target concept. The core marks of content-manipulable representation, as specified in Section 2.1, are target-directedness, content-sensitive downstream use, and the possibility of system-level misrepresentation. None of these requires sophisticated inference or compositionality. The paradigmatic manifestations, including rule-governed inference and compositional recombination, are listed as explanatory marks of the kind of state the core marks define, not as additional threshold conditions a system must satisfy before its states qualify.

The deeper point is this: the paper does not define the target and then charge existing theories with failing to explain it. It identifies the target as the kind of state that the major theories of cognition themselves are committed to explaining. LOT posits mental symbols that support inferential operations because of their content. Teleosemantics explains what makes a state have the content it has, presupposing a state that has content at all. Predictive processing, in its representational reading, attributes to the system a generative model whose states bear structured expectations about the world. The target is what these theories presuppose and in many cases aim to illuminate. The genesis question asks how any system first comes to have states of this kind, and the diagnosis is that when these theories are asked to answer that question, they begin downstream of the answer.

The objection would have force if the paper were arguing that existing theories fail to explain something they do not claim to explain. But the argument is different: it is that the explanatory primitives these theories deploy, when pressed into genesis service, already describe systems that have crossed the transition whose crossing is what genesis must explain. This is a claim about the starting point of these theories relative to the genesis problem, not a claim that they set their targets wrongly for their own purposes.

\subsection{The Hybrid-Theory Objection}

Even if no single framework can explain genesis, a hybrid theory combining information theory, teleosemantics, developmental biology, and mechanistic neuroscience might succeed where individual frameworks fail. The argument establishes at most that no current theory individually resolves genesis; it does not show that a suitably constructed hybrid cannot do so. On this reading, the diagnosis motivates theoretical development rather than the pessimistic conclusion that the existing vocabulary is categorically inadequate.

The paper does not claim that hybrid theories cannot in principle answer the genesis question. The claim is more specific: a hybrid theory will avoid the diagnosis only if each of its components satisfies the CSIDE condition with respect to genesis. Hybridity as such does not constitute an escape.

To see why, consider how a hybrid theory of genesis would be structured. It would invoke natural information relations to account for the informational resources available to the pre-genesis system, selectional mechanisms to account for the developmental or evolutionary trajectory of the system, and mechanistic organization to account for the physical substrate of representational states. But the diagnosis applies at the point where any of these components is invoked to explain not just what conditions are present in the pre-genesis system but why those conditions constitute or give rise to content-manipulable representational organization. At that point, the hybrid theory must either invoke concepts that are CSIDE in the genesis context, generating the same pattern of deferral identified in the individual frameworks, or it must do something that none of the existing individual frameworks does: describe the transition using conceptual resources that satisfy CC-1 and CC-2. If it does the latter, it is no longer drawing on the existing categorical vocabulary in the relevant sense, and its success would vindicate the higher-order inquiry into category adequacy rather than refuting it.

The hybrid-theory objection is therefore not a counter-example to the diagnosis but a specification of what a successful response to it would have to achieve. The paper welcomes hybrid theoretical development; what it resists is the assumption that aggregating existing frameworks whose individual components are subject to the presupposition diagnosis will, by the mere fact of aggregation, generate an account that escapes the diagnosis. Aggregating deferred resources does not resolve the deferral.

\subsection{The Empirical Theory Objection}

Biology and cognitive science already explain how representational systems arise through evolution and development\footnote{Carey, S. (2009). The Origin of Concepts. Oxford University Press, chs. 2-4. Core knowledge systems are the starting point for conceptual development; the genesis of those systems themselves is not the explanatory target.}\footnote{Spelke, E. S., and Kinzler, K. D. (2007). Core knowledge. Developmental Science, 10(1), 89-96.}\footnote{Margolis, E., and Laurence, S. (1999). Concepts: Core Readings. MIT Press. Margolis, E., and Laurence, S. (2007). The ontology of concepts. Philosophy Compass, 2(1), 80-91.}. Carey's work on the origin of concepts, Spelke's core knowledge program, and the broader literature on conceptual development collectively trace how representational repertoires emerge and expand. Why, then, is a philosophical diagnosis needed? The genesis problem may simply be an empirical problem awaiting empirical resolution.

Reply: The empirical accounts in cognitive development and evolutionary biology do not escape the presupposition structure; they relocate it. Carey's core knowledge account begins with innate representational systems as its starting point. Spelke's core knowledge framework posits innate representational states whose content is evolutionarily specified. Burge's objectification process begins with sensory systems already capable of forming proto-representational states. All of these accounts explain how existing representational systems develop and extend; they do not explain how a lineage of organisms first came to have systems with content-manipulable representational states.

Moreover, even a complete empirical history of how representational systems arose would leave open the philosophical question of what content-manipulable representation is such that it can arise by physical processes from non-representing substrates. That question is prior to the empirical history in the sense that the empirical history presupposes an answer to it. The higher-order inquiry into category adequacy is not a competitor to empirical investigation but a precondition for understanding what empirical investigation of genesis would have to establish.

\section{Conclusion}

The argument of this paper has three layers, each resting on the one before it, and they should be evaluated in that order.

The first and most securely established is the diagnosis itself. When the major frameworks in philosophy of mind are examined with respect to the genesis question rather than their intended explanatory targets, a recurring structural feature becomes visible: at some identifiable step, each framework deploys a concept whose explanatory purchase depends on the system under analysis already having content-manipulable representational organization. The frameworks differ in every other respect, in their ontological commitments, their methodologies, their accounts of what content is and how it is fixed. What they share is where they begin, relative to genesis. The case analyses in Section 3 constitute the primary evidence for this result, and the analytical template provided there is available to any reader who wishes to test whether a framework not examined here exhibits the same structure.

The second result follows from the first. If every attempt to explain genesis within the existing categorical vocabulary faces the presupposition structure, then the vocabulary itself generates systematic explanatory deferral: each proposed grounding step imports cognition-side resources, and any attempt to explain how those resources are themselves present in the pre-representational system encounters the same structure at the next level. This is what the paper calls the Representation Regress. The claim is not that genesis is inexplicable in principle, nor that the frameworks are internally incoherent. It is that the existing vocabulary defers rather than resolves the problem, and does so systematically.

The third result is the most tentative and should be read as such. The pattern of explanatory deferral motivates a question about the conceptual tools available for genesis explanation: whether the categories that philosophy of mind has developed are adequate for this specific explanatory task, or whether genesis requires conceptual resources that first-order theorizing cannot supply from within itself. This paper does not construct such resources. What it does is establish that the space for such an inquiry exists, and that two minimum conditions on any theory that occupies it can be derived: the Non-Presupposition Condition and the Transition Intelligibility Condition set out in Section 5. We have used the label meta-ontology of mind as a provisional shorthand for this space, without foreclosing questions about its ultimate disciplinary character. What is not provisional is the diagnostic result: the genesis problem is real, it is prior to the problems that existing frameworks are equipped to address, and the appearance of large language models has made it impossible to treat it as a merely theoretical concern.

The frameworks examined here recurrently presuppose elements of the target side of the transition that representation genesis is supposed to explain. To identify this pattern with precision, and to specify the conditions that any framework aiming to avoid it would need to satisfy, is the primary contribution of this paper. The question of how content-manipulable representation first arises from a non-representing physical substrate is foundational for any theory that seeks to explain representation rather than merely presuppose it. Recognizing this pattern is not a counsel of despair; it is an attempt to identify exactly where the explanatory work remains to be done.

\textbf{References}

Bechtel, W. (2008). Mental Mechanisms: Philosophical Perspectives on Cognitive Neuroscience. Routledge.

Brandom, R. (1994). Making It Explicit: Reasoning, Representing, and Discursive Commitment. Harvard University Press.

Brentano, F. (1874/1995). Psychology from an Empirical Standpoint (A. C. Rancurello, D. B. Terrell, and L. L. McAlister, Trans.). Routledge.

Bubeck, S., Chandrasekaran, V., Eldan, R., Gehrke, J., Horvitz, E., Kamar, E., Lee, P., Lee, Y. T., Li, Y., Lundberg, S., Nori, H., Palangi, H., Ribeiro, M. T., and Zhang, Y. (2023). Sparks of artificial general intelligence: Early experiments with GPT-4. arXiv preprint arXiv:2303.12712.

Burge, T. (2010). Origins of Objectivity. Oxford University Press.

Carey, S. (2009). The Origin of Concepts. Oxford University Press.

Chalmers, D. J. (1996). The Conscious Mind: In Search of a Fundamental Theory. Oxford University Press.

Clark, A. (2016). Surfing Uncertainty: Prediction, Action, and the Embodied Mind. Oxford University Press.

Clark, A., and Toribio, J. (1994). Doing without representing? Synthese, 101(3), 401-431.

Dretske, F. (1981). Knowledge and the Flow of Information. MIT Press.

Dretske, F. (1988). Explaining Behavior: Reasons in a World of Causes. MIT Press.

Egan, F. (2014). How to think about mental content. Philosophical Studies, 170(1), 115-135.

Fedorenko, E., Piantadosi, S. T., and Gibson, E. A. (2024). Language is primarily a tool for communication rather than thought. Nature, 630(8017), 575-586.

Fine, K. (2012). Guide to ground. In F. Correia and B. Schnieder (Eds.), Metaphysical Grounding: Understanding the Structure of Reality (pp. 37-80). Cambridge University Press.

Fodor, J. A. (1975). The Language of Thought. Harvard University Press.

Fodor, J. A. (1981). Representations: Philosophical Essays on the Foundations of Cognitive Science. MIT Press.

Fodor, J. A. (1998). Concepts: Where Cognitive Science Went Wrong. Oxford University Press.

Fodor, J. A. (2008). LOT 2: The Language of Thought Revisited. Oxford University Press.

Friston, K. (2010). The free-energy principle: A unified brain theory? Nature Reviews Neuroscience, 11(2), 127-136.

Friston, K., and Stephan, K. E. (2007). Free-energy and the brain. Synthese, 159(3), 417-458.

Gładziejewski, P. (2016). Predictive coding and representationalism. Synthese, 193(2), 559-582.

Gładziejewski, P., and Miłkowski, M. (2017). Structural representations: Causally relevant and different from detectors. Biology and Philosophy, 32(3), 337-355.

Harnad, S. (1990). The symbol grounding problem. Physica D: Nonlinear Phenomena, 42(1-3), 335-346.

Hempel, C. G., and Oppenheim, P. (1948). Studies in the logic of explanation. Philosophy of Science, 15(2), 135-175.

Hohwy, J. (2013). The Predictive Mind. Oxford University Press.

Husserl, E. (1900-1901/1970). Logical Investigations (2 vols., J. N. Findlay, Trans.). Routledge.

Husserl, E. (1939/1973). Experience and Judgment (J. S. Churchill and K. Ameriks, Trans.). Northwestern University Press.

Husserl, E. (1966/2001). Analyses Concerning Passive and Active Synthesis (A. Steinbock, Trans.). Kluwer.

Hutto, D. D., and Myin, E. (2013). Radicalizing Enactivism: Basic Minds without Content. MIT Press.

Kant, I. (1781/1787/1998). Critique of Pure Reason (P. Guyer and A. W. Wood, Trans.). Cambridge University Press.

Kripke, S. A. (1980). Naming and Necessity. Harvard University Press.

Lake, B. M., and Murphy, G. L. (2023). Word meaning in minds and machines. Psychological Review, 130(2), 401-431.

Levine, J. (1983). Materialism and qualia: The explanatory gap. Pacific Philosophical Quarterly, 64(4), 354-361.

Mahowald, K., Ivanova, A. A., Blank, I. A., Kanwisher, N., Tenenbaum, J. B., and Fedorenko, E. (2024). Dissociating language and thought in large language models: A cognitive perspective. Trends in Cognitive Sciences, 28(6), 517-540.

Margolis, E., and Laurence, S. (1999). Concepts: Core Readings. MIT Press.

Margolis, E., and Laurence, S. (2007). The ontology of concepts. Philosophy Compass, 2(1), 80-91.

McDowell, J. (1994). Mind and World. Harvard University Press.

Millikan, R. G. (1984). Language, Thought, and Other Biological Categories. MIT Press.

Millikan, R. G. (2004). Varieties of Meaning. MIT Press.

Mondorf, P., and Plank, B. (2024). Beyond accuracy: Evaluating the reasoning behavior of large language models, a survey. arXiv preprint arXiv:2404.01869.

Neander, K. (2017). A Mark of the Mental: In Defense of Informational Teleosemantics. MIT Press.

Papineau, D. (1987). Reality and Representation. Blackwell.

Papineau, D. (1993). Philosophical Naturalism. Blackwell.

Peacocke, C. (1992). A Study of Concepts. MIT Press.

Pearl, J. (2009). Causality: Models, Reasoning, and Inference (2nd ed.). Cambridge University Press.

Piccinini, G. (2020). Neurocognitive Mechanisms: Explaining Biological Cognition. Oxford University Press.

Putnam, H. (1975). The meaning of "meaning." In Mind, Language and Reality: Philosophical Papers, Volume 2 (pp. 215-271). Cambridge University Press.

Searle, J. R. (1980). Minds, brains, and programs. Behavioral and Brain Sciences, 3(3), 417-424.

Sellars, W. (1956). Empiricism and the philosophy of mind. Minnesota Studies in the Philosophy of Science, 1, 253-329.

Ramsey, W. M. (2007). Representation Reconsidered. Cambridge University Press.

Shannon, C. E. (1948). A mathematical theory of communication. Bell System Technical Journal, 27(3), 379-423.

Shea, N. (2018). Representation in Cognitive Science. Oxford University Press.

Sider, T. (2011). Writing the Book of the World. Oxford University Press.

Spelke, E. S., and Kinzler, K. D. (2007). Core knowledge. Developmental Science, 10(1), 89-96.

Thompson, E. (2007). Mind in Life: Biology, Phenomenology, and the Sciences of Mind. Harvard University Press.

van Inwagen, P. (1998). Meta-ontology. Erkenntnis, 48(2-3), 233-250.

Wan, Y., Wang, W., Yang, Y., Yuan, Y., Huang, J. T., He, P., Li, H., and Lyu, M. R. (2024). A and B equals B and A: Triggering logical reasoning failures in large language models. arXiv preprint arXiv:2401.00757.

Zahavi, D. (2003). Husserl's Phenomenology. Stanford University Press.

Zhou, B., Zhang, H., Chen, S., et al. (2024). Conceptual and unbiased reasoning in language models. arXiv preprint arXiv:2404.00205.

\end{document}